\documentclass{article}

\usepackage{PRIMEarxiv}

\usepackage[utf8]{inputenc} % allow utf-8 input
\usepackage[T1]{fontenc}    % use 8-bit T1 fonts
\usepackage{hyperref}       % hyperlinks
\usepackage{url}            % simple URL typesetting
\usepackage{booktabs}       % professional-quality tables
\usepackage{amsfonts}       % blackboard math symbols
\usepackage{nicefrac}       % compact symbols for 1/2, etc.
\usepackage{microtype}      % microtypography
\usepackage{lipsum}
\usepackage{fancyhdr}       % header
\usepackage{graphicx}       % graphics
\graphicspath{{media/}}     % organize your images and other figures under media/ folder
\usepackage{amsmath}
\usepackage{multirow}

\title{KITE-DDI: A Knowledge graph Integrated Transformer Model for accurately predicting Drug-Drug Interaction events from Drug SMILES and Biomedical Knowledge graph
\thanks{This work has been submitted to the IEEE for possible publication. Copyright may be transferred without notice, after which this version may no longer be accessible.} 
}

\author{
  Azwad Tamir \\
  Department of ECE \\
  University of Central Florida \\
  Orlando\\
  \texttt{azwad.tamir@ucf.edu} \\
   \And
  Jiann-Shiun Yuan \\
  Department of ECE \\
  University of Central Florida \\
  Orlando\\
  \texttt{jiann-shiun.Yuan@ucf.edu} \\
}

\begin{document}
\maketitle

\begin{abstract}
It is a common practice in modern medicine to prescribe multiple medications simultaneously to treat diseases. However, these medications could have adverse reactions between them, known as Drug-Drug Interactions (DDI), which have the potential to cause significant bodily injury and could even be fatal. Hence, it is essential to identify all the DDI events before prescribing multiple drugs to a patient. Most contemporary research for predicting DDI events relies on either information from Biomedical Knowledge graphs (KG) or drug SMILES, with very few managing to merge data from both to make predictions. While others use heuristic algorithms to extract features from SMILES and KGs, which are then fed into a Deep Learning framework to generate output. In this study, we propose a KG-integrated Transformer architecture to generate an end-to-end fully automated Machine Learning pipeline for predicting DDI events with high accuracy. The algorithm takes full-scale molecular SMILES sequences of a pair of drugs and a biomedical KG as input and predicts the interaction between the two drugs with high precision. The results show superior performance in two different benchmark datasets compared to existing state-of-the-art models especially when the test and training sets contain distinct sets of drug molecules. This demonstrates the strong generalization of the proposed model, indicating its potential for DDI event prediction for newly developed drugs. The model does not depend on heuristic models for generating embeddings and has a minimal number of hyperparameters, making it easy to use while demonstrating outstanding performance in low-data scenarios.
% \lipsum[1]
\end{abstract}

% keywords can be removed

\section{Introduction}
Polypharmacy, the routine prescription of several medications concurrently, has become a prevalent practice in contemporary medicine\cite{DDIFetal}. This technique of treating patients with multiple medications has become more common with the rapid increase of the number of approved drugs commercially available for consumption, especially to elderly individuals or patients suffering from long-lasting medical ailments like diabetes, Cancer, chronic heart conditions, etc. When multiple drugs are taken simultaneously in clinical practice, it can make the treatment process more complicated and potentially lead to unexpected interactions between the drugs, which is known as Drug-Drug Interaction (DDI). In some cases, these interactions can even be fetal\cite{Zhang}. 

As a result, a physician needs to examine the DDI events between the newly prescribed drugs and all other drugs currently taken by the patient to make sure that no adverse reactions are occurring between them. This, in turn, makes it necessary to maintain comprehensive DDI databases of all approved drugs currently in circulation and also to investigate the interactions between newly developed drugs with all other already approved drugs. The current method of determining new DDI events is via wet lab testing by domain experts. This is a very time-consuming and resource-intensive process and one of the major steps in a drug approval pipeline, resulting in the elongation of the overall system and also making new drugs more expensive to the general public.

To mitigate this, there have been a lot of studies focused on computational methods in order to construct algorithmic models for predicting DDI events. These computational methods have shown comparable performance to traditional wet assays based in vivo and in vitro trials while requiring much less time and resources. Recent efforts towards DDI research can be categorized into four main groups: literature extraction-based, matrix factorization-based, ensemble learning-based, and network-based.

The literature extraction-based studies are concerned with implementing natural language processing (NLP) techniques and machine learning models to extract drug-drug interactions from the biomedical literature. Given the large number of research articles and publications that are continually being published, there is a significant amount of drug knowledge that could be extracted from these materials \cite{Liu2016-yo, Shen2018-vp, 10.1093/bioinformatics/btx659}. Hence, numerous deep learning-based NLP algorithms have been developed that could extract and identify drug-drug interactions (DDI) from textual data. Recent examples of such study include Sun et al.\cite{Sun2018-kv}, who devised a deep convolutional neural network based model called DCNN, which utilized multiple layers and small convolutions, to extract drug-drug interactions (DDIs). In a subsequent study\cite{e21010037}, the same team developed a Recurrent Hybrid Convolutional Neural Network model known as RHCNN, which incorporated both canonical convolution and dilated convolution to predict DDI events with good accuracy. Nevertheless, conventional CNNs are unable to address the issue of lengthy sentences as they solely focus on neighboring words and disregard long-term dependencies and deep patterns within the textual data. In an attempt to solve this issue, Kavuluru et al.\cite{Kavuluru2017-je} put forward a character-level Recurrent Neural Network model known as char-RNN as a means to tackle the long-term word dependency in the DDI extraction task. Although these approaches have shown promise, the task often necessitates meticulous human annotations, which can be quite time-consuming and resource-intensive.

The second group of studies concerning DDI research uses matrix representation and adjacency matrix to analyze the relationship between the interacting drugs in a DDI event. Much of the recent research in this area has used a technique known as Neural Collaborative Filtering which was first introduced by Xiangnan et al \cite{He2017-sd}. Here, they have improved the simple linear inner product operation which is used in case of traditional matrix factorization (MF) systems and replaced it with a neural network-based collaborative filtering system which can successfully extract and analyze the complex structure present in the network data. Some examples of recent scholarly works in this field include Yu et al.\cite{YU2022104098}, who have approached the identification of potential side effects of drugs as a matrix reconstruction task, employing a linear neural network layer. Other related works consist of a study by Shi et al.\cite{Shi2016-ew} who have introduced a model called TMFUF, which is based on triple matrix factorization to generate relationships between the different drugs in the dataset, several novel matrix factorization-based models were developed which is based on manifold learning and neural networks\cite{10.1093/bioinformatics/btaa501}. In addition, Zhu et al.\cite{9310317} developed a machine learning based network to learn representations of drug dependency and introduced a supervised learning technique called probabilistic dependent matrix tri-factorization (PDMTF) to predict adverse drug-drug interactions (ADDI). Furthermore, Yu et al.\cite{Yu2018-gv} proposed a novel algorithm called DDINMF, leveraging the principles of semi-nonnegative matrix factorization. Lastly, Zhang et al.\cite{ZHANG201890} put forward the manifold regularized matrix factorization method to predict DDI events with high accuracy.

The third group of research in this area consists of ensemble learning-based methods which combine multiple different algorithms called sub-models to generate the final output. The different sub-models are responsible for understanding various relationships between the input data, enabling it to generate better results when all the sub-models are put together compared to their individual performance. Some notable studies in this area include a semi-supervised deep learning architecture proposed by Deepika et al.\cite{DEEPIKA2018136}, who have used network representation learning and meta-learning on four different drug datasets to make DDI predictions using an ensemble architecture, other studies include Zhang et al.\cite{ZHANG2019189}, who introduced another ensemble learning based model called SFLLN where they have applied linear neighborhood regularization techniques for predicting DDI events, Zhu et al.\cite{9653822}, who implemented unified multi-attribute discriminative representation learning (MADRL) to device an ensemble model for Adverse DDI prediction which can extract both intrinsic and extrinsic features and connections between pair of drug molecules. Finally, Schwarz et al.\cite{Schwarz2021-io} proposed a Siamese multimodal neural network based model called AttentionDDI which uses the self-attention mechanism to combine various drug features like targets, pathways, gene expression profiles, etc into a latent vector that is then used to predict DDI events.  

The fourth and last research area includes network-based approaches where, the algorithms extract information from graphs and networks describing drug data. Many contemporary studies employ a cheminformatics library known as RDKit\cite{rdkit} to convert drug SMILES (Simplified Molecular Input Line Entry System) into molecular graphs that represent the chemical structure of the drug compounds using heuristic algorithms. RDkit could also be used to extract features from drug compounds with the help of domain expertise and available literature. Notable contemporary research in this area includes the work by Hu et al.\cite{Hu*2020Strategies} who focused on the development of different molecular learning techniques using pre-trained GNNs. They aimed to extract the chemical structure embedding of drug molecules. Chen et al.\cite{10.1093/bioinformatics/btab169} utilized a pretrained message-passing neural network to create structural representations of drugs. The input data for this network consisted solely of the number and chirality of atoms, as well as the type and orientation of bonds. Nyamabo et al.\cite{10.1093/bib/bbab441} developed a gated message-passing neural network (GMPNN-CS) to analyze the molecular graph representations of drugs. This network is capable of learning chemical substructures of varying sizes and shapes. The main objective of their research was to predict potential drug-drug interactions between pairs of drugs. Other works include GNN-MolGNet\cite{10.1093/bib/bbab109}, which is a highly effective molecular graph model that has been extensively pre-trained at both the node and graph levels. This comprehensive training allows it to extract valuable chemical insights and generate easily understandable representations. Qian et al.\cite{10.1371/journal.pcbi.1007068} used feature similarity and feature selection techniques to construct a gradient boosting-based classifier in the rich biomedical network made up of several entities to expedite the process and get reliable prediction performance. Their goal was to enhance efficiency and ensure accurate prediction results. Next, the LR-GNN\cite{10.1093/bib/bbab513} paper introduced a propagation rule that captures the node embeddings of each GCN encoding layer to construct link representations (LRs). Further development was made with the proposed GCNMK\cite{10.1093/bib/bbab511} model which used various mechanisms associated with different types of DDI. Here, two graph convolution kernels are created by expanding and reducing the associated DDI network to make accurate predictions.  

Many of these models used a type of data known as the Knowledge Graph (KG) which is a powerful tool that allows for the representation of entity relationships. By integrating data from various sources, it creates a comprehensive biomedical information repository. Graph embedding methods, such as TransE\cite{NIPS2013_1cecc7a7} and ComplEx\cite{pmlr-v48-trouillon16} could be utilized to analyze the information present with these graph data. These methods aim to learn dense vectors for both relations and nodes in a low-dimensional space. The resulting vectors are then utilized in downstream biomedical tasks\cite{10.1093/bioinformatics/btz718, 10.1093/bib/bbaa243}. 

Many of the previous methods solely focus on predicting drug interactions without thoroughly examining the significance of these interactions or analyzing the type of interactions that are occurring in a specific situation and the impact they have on human metabolism. This brought about a rise in studies concentrating more on the type of interactions and their impact while predicting DDI events. Some of these include the work by Ryu et al.\cite{doi:10.1073/pnas.1803294115} who presented drug interactions as 86 metabolic events, which were described using human-readable sentences. The authors put forward a deep learning model called DeepDDI, which utilizes drug chemical structural information to accurately predict each DDI event. Furthermore, Deng et al.\cite{10.1093/bioinformatics/btaa501} put forward a multimodal deep learning framework called DDIMDL. This system effectively integrates various drug features to accurately predict 65 DDI-related events. Lin et al.\cite{10.1093/bib/bbab421} examined a dataset containing 100 drug-drug interaction (DDI) events sourced from the DrugBank database.  

However, most of these studies use a method of splitting training and test set DDI events called the transductive setting \cite{Lee2019-gw}. Here, the training and test sets consist of common drug compounds so the algorithm could predict DDI events by analyzing relationships and patterns within drug compounds with known action mechanisms. However, it is important to consider the inductive setting, also known as cold-start scenarios which most previous studies have overlooked\cite{10.1093/bioinformatics/btaa501, 10.1093/bib/bbab441, 10.1093/bib/bbab421, 10.1093/bib/bbab133}. In this setting, all available drugs are first divided into test and training sets. Later, the DDI events that exist between drug pairs in each split are gathered to make up the datapoints in the training and test datasets. This was identified in the MSEDDI model\cite{ijms24054500}, where a multi-classification DDI event prediction system was proposed which used the more challenging inductive dataset split technique to evaluate their results. It uses a multi-structural deep learning framework to combine information from multiple sources to achieve state-of-the-art performance for DDI prediction in the more challenging inductive setting and hence has been used as the primary benchmark method for performance comparison with the proposed model in this work. However, the MSEDDI model used various heuristic models that are based on domain expert knowledge to extract features from the SMILE notations of drug compounds to make DDI predictions. This makes it dependent on other models and cannot be implemented to generate an end-to-end pipeline that could input raw drug SMILES and output DDI classifications. 

In this work, we have proposed an end-to-end fully machine learning based model that only takes raw drug SMILEs and a biological knowledge graph as input and predicts DDI events with high accuracy. The model is evaluated on two different datasets and shows better performance compared to state-of-the-art models, especially in the more challenging inductive setting. This indicates that the proposed architecture is better at extracting the latent chemical features buried in the drug SMILEs and can also understand the deep structure and patterns responsible within the biomedical knowledge graphs that is required for making highly accurate predictions in the inductive setting.  The key contributions and novelty of this work are outlined below:

\begin{itemize}
  \item Superior performance and accuracy compared to other related state-of-the-art models especially on predictions in the inductive dataset split setting showing good generalization.
  \item It is an end-to-end machine learning model with no heuristic components that rely on domain expert knowledge.
  \item The proposed model is lightweight, computationally inexpensive, and easy to use as it is an end-to-end pipeline requiring very few hyperparameter optimizations.
  \item The proposed algorithm only requires 2 inputs (KG and SMILES) as opposed to the main benchmark model which requires five (KG, SMILES, MPNN, AFP, WEAVE)\cite{ijms24054500}. These other inputs need to be generated using separate full-scale algorithms raising the dependency of the method on other models.
  \item The proposed architecture shows better performance at low data settings compared to the main benchmark method.
\end{itemize}

The model architecture and details about the dataset including preprocessing steps and evaluation criteria are given in section 2. The experimental results, ablation study, and detailed analysis of the results are outlined in section 3. Finally, further discussion, overall outcome, and future research direction are given in section 4. 

% \lipsum[2]
% \lipsum[3]
\section{Methodology}
\subsection{Dataset}
Two different datasets have been prepared to evaluate the performance of KITE-DDI and compare it with other similar benchmark methods. The first dataset (Dataset 1) is prepared following the same process as Deng et al\cite{10.1093/bioinformatics/btaa501}. It consists of 572 drugs and 37,264 DDI events which were all sourced from the DrugBank repository\cite{DrugBank}. Each DDI event corresponds to a drug pair and represents an increase or decrease in the interaction between the two drugs on the metabolism level of the human body totalling to 65 events. The second dataset (Dataset 2) is created analogously to the paper by Lin et al\cite{10.1093/bib/bbab421}. Here, 1258 drugs were collected from the DragBank repository\cite{DrugBank} which consisted of 325,539 pairwise DDI event datapoints in total and comprised of 100 unique DDI classes.

\begin{figure}[h]
  \centering
  % \fbox{\rule[-.5cm]{4cm}{4cm} \rule[-.5cm]{4cm}{0cm}}
  \includegraphics[width=0.5\linewidth]{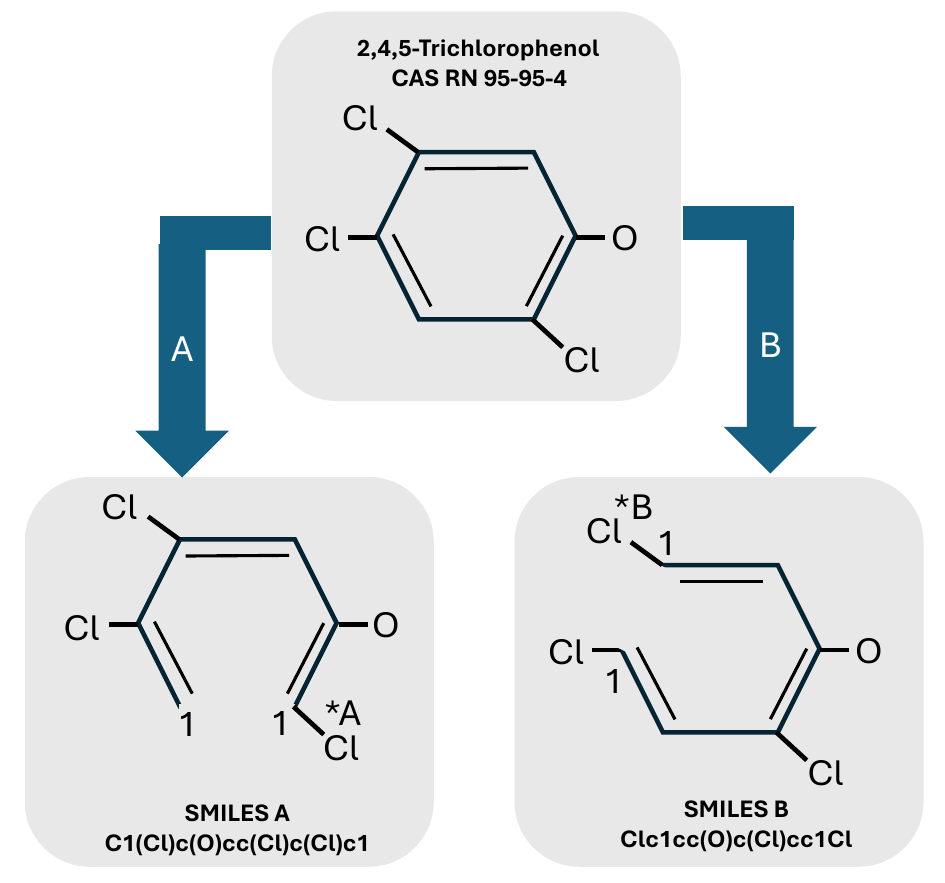}
  \caption{Multiple equally valid SMILES representation of a single compound.}
  % \Description{ROC curves}
  \label{Fig_SMILES}
\end{figure}

Each data point in the dataset consists of a pair of drugs along with their Simplified Molecular Input Line Entry System (SMILES) notation which is a representation of the three-dimensional chemical structure of the drug molecule with the help of a string of symbols. The SMILE notation for representing chemical compounds is a well-accepted and practiced method for inputting drug compounds into computational algorithms and machine learning models\cite{epa}. However, there are certain limitations to the SMILE notation due to its conversion of three-dimensional chemical compounds into a linear two-dimensional format. One of the major ones related to this study involves the representation of a single molecule by multiple equally valid but different SMILES notations. This happens in cyclic structures where different notations may arise depending on the starting position of the cyclic compound. This phenomenon is illustrated in Fig. \ref{Fig_SMILES}, which shows two different SMILE notations for the same cyclic molecule, “2,4,5-Trichlorophenol CAS RN 95-95-4". To resolve this problem, a randomizer augmentation is attached to the dataset module which feeds the data into the model. This randomizer generates a different but valid SMILES representation each time a drug is inserted into the model which teaches it about the existence and structure of the multiple representations of the same molecule.

%break

Apart from the drug SMILES, the other input of the proposed algorithm is a Biomedical Knowledge graph called the “Drug Repurposing Knowledge Graph” (DRKG). This graph consists of information relating to genes, compounds, biological pathways, diseases, proteins, side effects, symptoms, etc. It consists of about a hundred thousand nodes divided into 13 entry types and around 6 million edges belonging to 107 types. Previous research suggests that the biomedical knowledge graphs contain vital information about DDI events which could be leveraged to achieve better performance in predicting drug interactions. 

\begin{table}
\centering
\renewcommand{\arraystretch}{1.3}
\caption{\textbf{Number of Datapoints in each split of Dataset 1 and 2}}
\label{table_data}
\setlength{\tabcolsep}{10pt}
\begin{tabular}{|c||c|c|}
\hline
\textbf{Splits}                  & \textbf{Dataset 1} & \textbf{Dataset 2} \\ \hline \hline
Training set            & 20404     & 156713    \\ \hline
5-fold Cross-validation & 5101      & 39187     \\ \hline
U1 Split                & 10541     & 110828    \\ \hline
U2 Split                & 1080      & 15679     \\ \hline
Unique Drugs            & 572       & 1254      \\ \hline
\end{tabular}
\label{t_dataset}
\end{table}

Both Datasets 1 and 2 are randomly divided into five equal parts to create the fivefold cross-validation. In this process, one fold is fixed as the validation set while the other four folds are combined to create the training set. This step is repeated five times, changing the fold which is set as the validation, and the accuracy metrics averaged to evaluate the final results. These five-fold cross-validation results are used to measure the performance of the model in the transductive setting. To compute and compare the inductive performance of the model, two different dataset splits called U1 and U2 are also generated. In the case of U1, only one drug in each DDI pair in the test set is present in the training set while in the case of U2, both the drugs in each DDI pair in the test set are absent in the training set. As a result, the U1 dataset is more challenging compared to the fivefold cross-validation, while the U2 data split is the most challenging. The composition of training and testing DDI events in each dataset split is given in Table \ref{table_data}.  

\begin{table}
\centering
\renewcommand{\arraystretch}{1.3}
\caption{\textbf{Percentage composition of the most abundant DDI events in each Dataset and their description}}
\label{t_dist}
\setlength{\tabcolsep}{3pt}
\begin{tabular}{|l||l|l|}
\hline
\textbf{Events}                                                                                     & \textbf{Dataset 1 (\%)} & \textbf{Dataset 2 (\%)} \\ \hline
\begin{tabular}[c]{@{}l@{}}Drug A metabolism decreases when\\combined with Drug B\end{tabular}    & 25.76                  & 30.93                  \\ \hline
\begin{tabular}[c]{@{}l@{}}Risk or severity of adverse effects\\increases\end{tabular}
& 25.30                  & 10.89                  \\ \hline
Increase in serum concentration                                                                     & 15.80                  & 5.12                   \\ \hline
Decrease in serum concentration                                                                     & 4.94                   & 1.36                   \\ \hline
\begin{tabular}[c]{@{}l@{}}Therapeutic efficacy of Drug A\\decreases\end{tabular}                    & 3.48                   & 3.98                   \\ \hline
\begin{tabular}[c]{@{}l@{}}Excretion decreases resulting in\\decrease of serum level\end{tabular} & 0.33                   & 11.30                  \\ \hline
\begin{tabular}[c]{@{}l@{}}Drug A metabolism increases when\\combined with Drug B\end{tabular}    & 1.92                   & 8.16                   \\ \hline
\begin{tabular}[c]{@{}l@{}}Increase in the risk or severity of QTc \\ prolongation\end{tabular}     & 0.27                   & 5.35                   \\ \hline
\end{tabular}
\end{table}

In addition to these two datasets, an additional unlabeled dataset has also been prepared to pretrain the model in an unsupervised manner. The dataset consists of 17,741 drug SMILES sourced from the DurgBank \cite{DrugBank} online repository and the ChEMBL database \cite{ChEMBL}.  

The composition of the most abundant DDI events in the two datasets is given in Table \ref{t_dist}. It shows that these top four events account for about 75\% and 66\% in Dataset 1 and Dataset 2, respectively creating a long tail distribution with the least abundant events contributing to a small portion of the overall data leading to a skewed class representation.

\subsection{Transformer}
The transformer architecture was first introduced in the paper “Attention is all you need” 
\cite{attention_need}. It is a sequence-to-sequence model, meaning it takes a sequence as an input and outputs another. It started out as a natural language processing (NLP) algorithm but soon expanded to other tasks like computer vision, bioinformatics, and drug discovery. One of the major limitations of the transformer architecture is that it requires a large amount of data to be trained effectively and achieve optimal performance. One of the major components of the proposed algorithm in this work involves a transformer block but the architecture has been modified along with the training regime to adopt it for low data setting. 

The original transformer consisted of an encoder and a decoder. The encoder is responsible for taking the input sequence and converting it into a latent vector. The decoder, on the other hand, takes the previous token of the sequence along with the latent vectors coming from the encoder to generate the output sequence. Depending on the context of the application, several variants of the original transformer model were later proposed which could be divided into three major categories \cite{transformers_survey}:

\begin{itemize}
  \item Encoder Only: This type of architecture only consists of the encoder block and the decoder is absent \cite{ALBERT, BERT, DistilBERT, clark2020electra}. These are typically applied to tasks like sequence labeling, image classification, sequence ranking etc.
  \item Encoder-Decoder: The complete transformer architecture consists of both an encoder and a decoder module \cite{Bart, T5}. This is commonly used in applications like neural machine translation and sequence-to-sequence tasks.
  \item Decoder Only: These models only consist of a decoder module \cite{CTRL, GPT2, TransformerXL}. Applications of these types of models include sequence generation, language modeling, etc.
\end{itemize}

The transformer module used in this work is of the encoder-only variant. It consists of three encoders that take the tokenized input sequence and convert it into latent vectors. All the encoder modules consist of trainable weights which generate various features of the input sequence, that are then summed up to be fed into the encoder layers. The token embeddings contain information about the type of token that each sequence base consists of. The segment embeddings divide the input sequence into two segments for the pair of drug SMILES in each data point. These two embeddings are then combined to create the non-positional embedding vector. The final encoder is responsible for generating the positional embeddings, which also consist of learnable weights containing information about the position of each token in the input sequence. 

The encoder module is responsible for generating the representation of the input sequence in the form of a latent vector. It consists of attention heads which are the principal computational blocks in the transformer architecture followed by a normalization layer and a position-wise feedforward network before ending with another normalization layer. These layers make sure that the information flow does not become too large or too small which could potentially lead to the vanishing or exploding gradient problem. There is also a residual connection that bypasses the attention and feedforward network in each layer to make sure that the input information also reaches the later layers in the module. Multiple encoder layers are stacked on top of one another to create the overall architecture of the encoder module.

\subsection{Multiheaded Attention}
The basic component of an attention function involves the utilization of a query and a collection of key-value pairs to generate a corresponding output. All these elements are represented as vectors of dimension $d_k$ for queries and keys, and $d_v$ for values, respectively. The output is determined by computing a weighted sum of the values, where each weight is obtained from a compatibility function that assesses the degree of similarity between the query and the corresponding key. To obtain the weights for the values, a series of calculations are performed. Firstly, the dot product of the query with all the keys is computed. Then, each result is divided by the square root of $d_k$. Finally, a softmax function is applied to obtain the desired weights. For computation convenience, the set of queries is stacked into a matrix Q. Similarly, the keys and values are merged to form matrices K and V. The output matrix is then calculated according to \eqref{eq1}.

\begin{align*} Attention(Q,K,V) = softmax(\frac{QK^{T}}{\sqrt{d_{k}}})V\label{eq1}\tag{1}\end{align*}

Instead of utilizing a single attention function in which the keys, values, and queries are vectors with a dimension of $d_{model}$, it is advantageous to linearly project them h times into dimensions $d_k$, $d_k$, and $d_v$ respectively for the queries, keys and values using different learned linear projections. The attention function is subsequently executed simultaneously on the projections yielding output values of $d_v$ dimensions. The concatenated outputs are then projected once more to generate the final values. The utilization of multi-head attention allows the model to effectively incorporate information from diverse representation subspaces, enabling simultaneous attention at multiple positions. Utilizing just one attention head would result in averaging, potentially eliminating vital details. The computation of MultiHead attention is provided in \eqref{eq2} and \eqref{eq3} \cite{attention_need}. In this context, $W_{qi}$ and $W_{ik}$ have dimensions $d_{model}$ by $d_k$, while $W_{iv}$ has the shape $d_{model}$ by $d_v$. Due to the reduced dimension of each head, the computational cost of the overall operation is comparable to that of single-head attention with full dimensionality.

\begin{align*} MultiHead(Q,K,V) = concat(h_{1},h_{2}, ..., h_{h})W^{o}\label{eq2}\tag{2}\end{align*}

\begin{align*} h_{i} = Attention(QW_{q}^{i}, KW_{i}^{K}, VW_{i}^{V})\label{eq3}\tag{3}\end{align*}

\begin{figure}[]
  \centering
  % \fbox{\rule[-.5cm]{4cm}{4cm} \rule[-.5cm]{4cm}{0cm}}
  \includegraphics[width=\linewidth]{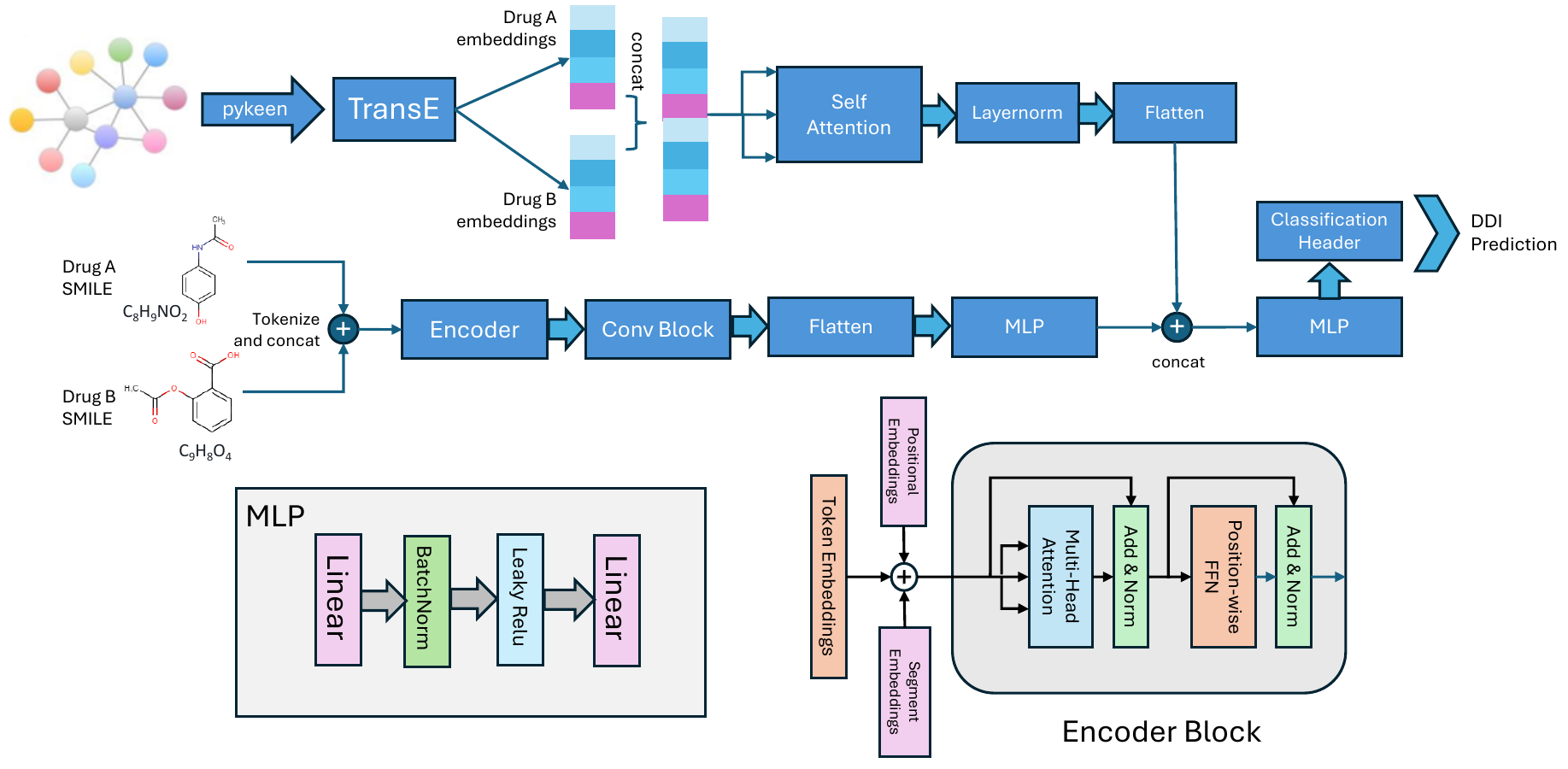}
  \caption{Overall architecture of the proposed KITE-DDI model. The internal layers of the MLP and encoder blocks are also illustrated in the figure.}
  % \Description{ROC curves}
  \label{Fig_model}
\end{figure}

% \Figure[t!](topskip=0pt, botskip=0pt, midskip=0pt)[width=1\linewidth]{Fig_Model_cropped.pdf}
% { \textbf{Overall architecture of the proposed KITE-DDI model. The internal layers of the MLP and encoder blocks are also illustrated in the figure.}\label{Fig_model}}

\subsection{Self-Attention}
The multiheaded attention layer from the transformer architecture could be used elsewhere with appropriate modifications. Self-attention is an effective and computationally efficient way of extracting long and short term patterns within a one-dimensional linear vector. This can be viewed as a fully connected layer where the weights are dynamically generated according to the pairwise input relationships. In this case, the input vector is copied three times and fed into the multiheaded attention mechanism as query, keys, and values. The output vector generated is the same shape as the input vector but contains information about the structure, dependencies, and patterns within the input. In the proposed model, the self-attention module is used to extract interaction features from the concatenated knowledge graph embeddings of the two drug compounds in the DDI event.

\begin{table}
\centering
\renewcommand{\arraystretch}{1.3}
\caption{\textbf{Computational complexity of Self-Attention compared to other algorithms}}
\label{t_comp}
\setlength{\tabcolsep}{10pt}
\begin{tabular}{|l||c|c|c|}
\hline
\textbf{Layer Type}         & \textbf{Complexity} & \begin{tabular}[c]{@{}c@{}}\textbf{Sequential}\\ \textbf{Operations}\end{tabular} & \begin{tabular}[c]{@{}c@{}}\textbf{Maximum}\\ \textbf{Path length}\end{tabular} \\ \hline \hline
Self-Attention              & O(n2 . d)           & O(1)                                                            & O(1)                                                           \\ \hline
Recurrent                   & O(n . d2)           & O(n)                                                            & O(n)                                                           \\ \hline
Convolutional               & O(k . n . d2)       & O(1)                                                            & O(logk (n))                                                    \\ \hline
Self-Attention (restricted) & O(r . n . d)        & O(1)                                                            & O(n/r)                                                         \\ \hline
\end{tabular}
\end{table}

Table \ref{t_comp} shows the comparison of the complexity, sequential operations, and maximum path length between self-attention and three frequently employed layer types. The key benefits of the self-attention layer are outlined below \cite{transformers_survey}: 

\begin{itemize}
  \item The maximum path length of this model is similar to that of fully connected layers, which makes it ideal for computing long-range dependencies. In contrast to fully connected layers, this approach demonstrates greater efficiency in terms of parameters and excels in handling inputs of varying lengths.
  \item Convolutional layers possess a restricted receptive field, often necessitating a deep network stack to attain a comprehensive global pattern extraction. On the other hand, the self-attention's constant maximum path length enables it to effectively capture long-range dependencies across various layers.
  \item The consistent number of sequential operations and the maximum path length contribute to the parallelizability and effectiveness of self-attention in long-range modeling, surpassing that of recurrent layers.
\end{itemize}

It could be observed from Table 3 that the self-attention layer understands interactions between all tokens through a consistent number of sequential operations in contrast to a recurrent layer which requires O(n) sequential operations. When the sequence length is shorter than the representation dimensionality, self-attention layers tend to be faster than recurrent layers. This is particularly true for sentence representations used in advanced machine translation models, like word-piece and byte-pair representations. In order to optimize computational performance for tasks that deal with lengthy sequences, it is possible to restrict self-attention to a specific neighborhood around each output position in the input sequence. This approach would result in an increase in the highest possible path length to O(n/r).

\subsection{Model}
The overall architecture of the proposed KITE-DDI model is illustrated in Fig. \ref{Fig_model}. The model has two inputs; the first one is from a biomedical Knowledge Graph (KG) and the second is from a pair of Drug SMILEs which are involved in the DDI event. The Drug Repurposing Knowledge Graph (DRKG) is used as the biomedical KG, which is then fed into the Pykeen deep learning framework\cite{pykeen} to extract the entity features using the TransE\cite{NIPS2013_1cecc7a7} algorithm. These embeddings for the two drug compounds are then concatenated into a one-dimensional vector of length 800 followed by a self-attention layer which extracts the information relating to the interaction between the two drug embeddings. 

On the other side, the SMILE representation of the two Drug compounds in the DDI event is randomized, tokenized, and concatenated into a single one-dimensional sequence which is then padded or concatenated to a fixed predefined dimension of 500 tokens. The sequence then goes into the encoder module, made up of 6 encoder layers with 8 multiheaded attention blocks. A model dimensionality of 256 is picked which seems to maximize efficiency while keeping the computational complexity relatively low. The length of the feedforward network is also kept equal to the model dimensionality at 256 to ensure stability. The output from the encoder layers are latent vectors of each token in the input sequence that are gathered into a multidimensional vector of shape 500 by 256. This latent vector then goes through a convolutional block to extract the features and shrink the dimension.

\begin{figure}
  \centering
  % \fbox{\rule[-.5cm]{4cm}{4cm} \rule[-.5cm]{4cm}{0cm}}
  \includegraphics[width=0.4\linewidth]{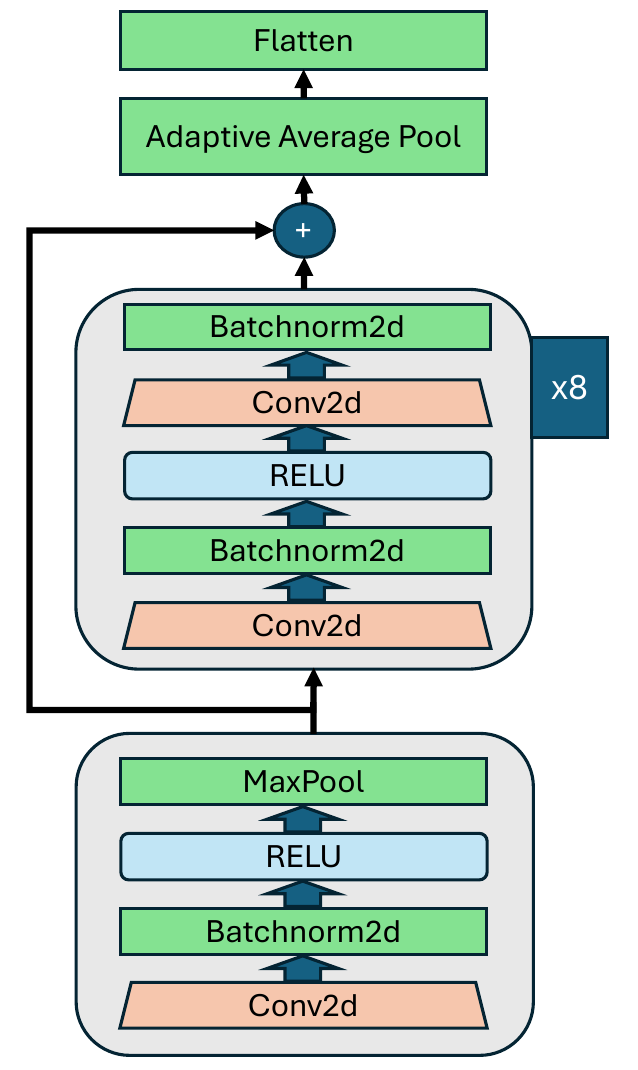}
  \caption{Block diagram of the convolutional module used in the proposed model.}
  % \Description{ROC curves}
  \label{Fig_conv}
\end{figure}

% \Figure[t!](topskip=0pt, botskip=0pt, midskip=0pt)[width=0.7\linewidth]{Fig_conv_cropped.pdf}
% { \textbf{Block diagram of the convolutional module used in the proposed model.}\label{Fig_conv}}

The structure of the convolutional module is illustrated in Fig. \ref{Fig_conv}. A single repeating block consists of a convolutional layer followed by a batch normalization layer, Relu activation, another convolutional layer, and a final batch normalization layer at the end. This block is then repeated 8 times with residual connections between each block. The extracted feature from this convolutional module is then flattened and fed into a Multilevel Perceptron (MLP) module consisting of a fully connected linear layer, batch normalization, leaky RELU activation function and another linear layer at the end. The output from the MLP is then concatenated with the KG embeddings and fed into another MLP module before outputting the output class predictions determined by a softmax layer.

% \usepackage{multirow}
\begin{table}[!t]
\centering
\renewcommand{\arraystretch}{1.6}
\caption{\textbf{Performance Comparison of proposed model with similar SOTA methods}}
\label{t_result1}
\setlength{\tabcolsep}{5pt}
\begin{tabular}{|c||c|cccc||cccc|}
\hline
\textbf{Dataset}                    & \textbf{Model}        & \multicolumn{4}{c||}{\textbf{U2 Split}}                                                                                                   & \multicolumn{4}{c|}{\textbf{U1 Split}}                                                                                                   \\ \hline \hline
\multirow{7}{*}{\textbf{Dataset 1}} & \textbf{Metrics}      & \multicolumn{1}{c|}{\textbf{Accuracy}} & \multicolumn{1}{c|}{\textbf{F1 Score}} & \multicolumn{1}{c|}{\textbf{AUPR}}   & \textbf{AUC}    & \multicolumn{1}{c|}{\textbf{Accuracy}} & \multicolumn{1}{c|}{\textbf{F1 Score}} & \multicolumn{1}{c|}{\textbf{AUPR}}   & \textbf{AUC}    \\ \cline{2-10} 
                                    & \textbf{MSEDDI}       & \multicolumn{1}{c|}{44.51\%}            & \multicolumn{1}{c|}{0.1691}            & \multicolumn{1}{c|}{0.3999}          & 0.9543          & \multicolumn{1}{c|}{65.17\%}            & \multicolumn{1}{c|}{0.4771}            & \multicolumn{1}{c|}{\textbf{0.6810}} & \textbf{0.9823} \\ \cline{2-10} 
                                    & \textbf{DeepDDI}      & \multicolumn{1}{c|}{36.02\%}            & \multicolumn{1}{c|}{0.1373}            & \multicolumn{1}{c|}{0.2781}          & 0.9059          & \multicolumn{1}{c|}{57.74\%}            & \multicolumn{1}{c|}{0.3416}            & \multicolumn{1}{c|}{0.5594}          & 0.9575          \\ \cline{2-10} 
                                    & \textbf{Lee's method} & \multicolumn{1}{c|}{40.97\%}            & \multicolumn{1}{c|}{0.2022}            & \multicolumn{1}{c|}{0.3184}          & 0.8302          & \multicolumn{1}{c|}{64.05\%}            & \multicolumn{1}{c|}{0.5039}            & \multicolumn{1}{c|}{0.6244}          & 0.9247          \\ \cline{2-10} 
                                    & \textbf{DDIMDL}       & \multicolumn{1}{c|}{40.75\%}            & \multicolumn{1}{c|}{0.1590}            & \multicolumn{1}{c|}{0.3635}          & 0.9512          & \multicolumn{1}{c|}{64.15\%}            & \multicolumn{1}{c|}{0.4460}            & \multicolumn{1}{c|}{0.6558}          & 0.9799          \\ \cline{2-10} 
                                    & \textbf{MSD-SA-DDI}   & \multicolumn{1}{c|}{43.78\%}            & \multicolumn{1}{c|}{0.2326}            & \multicolumn{1}{c|}{0.3810}          & 0.8675          & \multicolumn{1}{c|}{64.59\%}            & \multicolumn{1}{c|}{0.5471}            & \multicolumn{1}{c|}{0.6390}          & 0.9435          \\ \cline{2-10} 
                                    & \textbf{KITE-DDI}     & \multicolumn{1}{c|}{\textbf{51.29\%}}   & \multicolumn{1}{c|}{\textbf{0.4698}}   & \multicolumn{1}{c|}{\textbf{0.4648}} & \textbf{0.9649} & \multicolumn{1}{c|}{\textbf{67.21\%}}   & \multicolumn{1}{c|}{\textbf{0.6628}}   & \multicolumn{1}{c|}{0.6727}          & 0.9729          \\ \hline \hline
\multirow{6}{*}{\textbf{Dataset 2}} & \textbf{MSEDDI}       & \multicolumn{1}{c|}{63.09\%}            & \multicolumn{1}{c|}{0.3111}            & \multicolumn{1}{c|}{0.6596}          & 0.9863          & \multicolumn{1}{c|}{76.97\%}            & \multicolumn{1}{c|}{0.6486}            & \multicolumn{1}{c|}{0.8315}          & 0.9947          \\ \cline{2-10} 
                                    & \textbf{DeepDDI}      & \multicolumn{1}{c|}{36.11\%}            & \multicolumn{1}{c|}{0.1868}            & \multicolumn{1}{c|}{0.2820}          & 0.9264          & \multicolumn{1}{c|}{58.83\%}            & \multicolumn{1}{c|}{0.4709}            & \multicolumn{1}{c|}{0.5851}          & 0.9746          \\ \cline{2-10} 
                                    & \textbf{Lee's Method} & \multicolumn{1}{c|}{48.67\%}            & \multicolumn{1}{c|}{0.3082}            & \multicolumn{1}{c|}{0.4349}          & 0.9093          & \multicolumn{1}{c|}{69.17\%}            & \multicolumn{1}{c|}{0.5934}            & \multicolumn{1}{c|}{0.7119}          & 0.9687          \\ \cline{2-10} 
                                    & \textbf{DDIMDL}       & \multicolumn{1}{c|}{46.99\%}            & \multicolumn{1}{c|}{0.3032}            & \multicolumn{1}{c|}{0.4386}          & 0.9685          & \multicolumn{1}{c|}{67.20\%}            & \multicolumn{1}{c|}{0.5817}            & \multicolumn{1}{c|}{0.7086}          & 0.9885          \\ \cline{2-10} 
                                    & \textbf{MSD-SA-DDI}   & \multicolumn{1}{c|}{47.94\%}            & \multicolumn{1}{c|}{0.2937}            & \multicolumn{1}{c|}{0.4450}          & 0.9686          & \multicolumn{1}{c|}{66.64\%}            & \multicolumn{1}{c|}{0.5919}            & \multicolumn{1}{c|}{0.6820}          & 0.9862          \\ \cline{2-10} 
                                    & \textbf{KITE-DDI}     & \multicolumn{1}{c|}{\textbf{66.96\%}}   & \multicolumn{1}{c|}{\textbf{0.6537}}   & \multicolumn{1}{c|}{\textbf{0.7197}} & \textbf{0.9886} & \multicolumn{1}{c|}{\textbf{79.09\%}}   & \multicolumn{1}{c|}{\textbf{0.7834}}   & \multicolumn{1}{c|}{\textbf{0.8528}} & \textbf{0.9948} \\ \hline
\end{tabular}
\end{table}

\subsection{Training}
The first step of the training involves pretraining the transformer encoder so that it understands the detailed structure of the incoming SMILES representation of the drug compounds. This is done with the help of the pretraining dataset consisting of 17,741 drug SMILES. For the first step of the process, the drug SMILES are converted into pairwise DDI datapoints. The first drug in a datapoint is selected chronologically from the dataset and it is paired up with another randomly selected SMILE. The two drug SMILES are then tokenized and concatenated to form a single sequence. A vocabulary dictionary is created from the pretrained SMILES for the tokenization process which is later reused in the finetuning step as well. The input sequence is truncated if it exceeds the maximum allowed sequence length of 500 tokens and padded if it is shorter to make all of them the same length.

\begin{figure}[h]
  \centering
  % \fbox{\rule[-.5cm]{4cm}{4cm} \rule[-.5cm]{4cm}{0cm}}
  \includegraphics[width=0.6\linewidth]{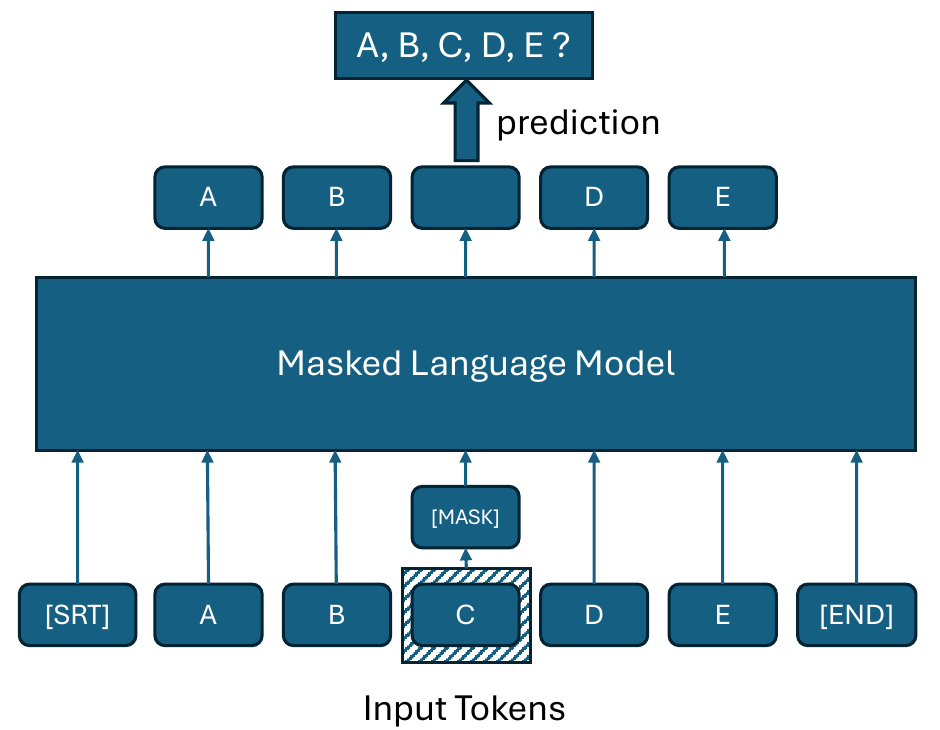}
  \caption{Masked Language Modelling Technique used for pretraining the Transformer encoder blocks.}
  % \Description{ROC curves}
  \label{Fig_MLM}
\end{figure}

% \Figure[t!](topskip=0pt, botskip=0pt, midskip=0pt)[width=0.9\linewidth]{Fig_MLM_cropped.pdf}
% { \textbf{Masked Language Modelling Technique used for pretraining the Transformer encoder blocks.}\label{Fig_MLM}}

A masked language modelling (MLM) technique is used to pretrain the transformer module in an unsupervised manner. Here, 15\% of the tokens in the input sequence are randomly masked using a special masking token and the task of the training algorithm is to correctly predict the hidden bases. The illustration of the overall MLM process is given in Fig \ref{Fig_MLM}. A cross-entropy loss function given in \ref{eq4} is used to calculate the loss in each step of the training process and the Adam optimizer is used with a learning rate of 1e-5. The model is pretrained for a total of 700 epochs with a batch size of 8.

\begin{align*} Loss = -\frac{1}{N} \sum_{i}^{N}\sum_{j}^{M} y_{ij}log(\hat{y}_{ij})\label{eq4}\tag{4}\end{align*}

After the pretraining is done, the learned weights for the encoder layer and the positional and non-positional embeddings are transferred to the main model which is then finetuned on the application datasets. The other modules apart from the transformer block are randomly initialized. The Cross-entropy loss function is used for the overall training along with an Adam optimizer with a learning rate of 5e-5 and a weight decay of 1e-5. The model is trained up to 100 epochs with a batch size of 32 on a single Nivida RTX 2070 Ti GPU requiring around 4 GB of memory and takes approximately 5 hours to train. The model is then evaluated on Dataset 1 and 2 on both the transductive and inductive settings to investigate the generalization and performance of the model compared to other recent state-of-the-art methods.

\section{Result}
\label{sec:result}
This section presents the performance of the proposed algorithm in DDI event prediction along with comparison with other state of the art models and methods. The performance of the proposed method, KITE-DDI, compared to other state of the art models in four different evaluation metrics are given in Table \ref{t_result1}. Accuracy, F1 score, AUPR, and AUC have been used as the different performance metrics. Here, the accuracy is given by the number of correctly labeled DDI events for a pair of drugs divided by the total number of samples. The F1 score is a combination of the precision and recall values for the predictions samples and is given in \eqref{eq5}. The AUPR denotes the area under the precision-recall curve. It is normalized to a maximum value of unity where a larger value corresponds to better performance. Lastly, the AUC computes the area under an ROC curve and, like the AUPR, ranges between 0 and 1.

\begin{align*} F1\:score = \frac{TP}{TP+\frac{1}{2}(FP+FN)}\label{eq5}\tag{5}\end{align*}

The results show that the proposed model is able to outperform the other methods in most metrics in both the Datasets. Here, U1 split refers to the inductive test split where one the drugs in each DDI pair is not present in the training set and has not been seen before by the algorithm. On the other hand, the U2 split consists of DDI pairs where none of the two drugs are present in the training set. The U1 and U2 splits create a more challenging task, where the algorithm must understand the latent processes in which the drug pairs interact to make DDI predictions. 

For Dataset 1, KITE-DDI outperformed MSEDDI, which is the principal benchmark method in this study, by 6.78\% and 2.04\% in accuracy in the U2 and U1 split tasks, respectively. It also managed to show better performance in terms of accuracy, F1 score, AUPR and AUC compared to the other algorithms in the U2 split and in accuracy and F1 score for the U1 split. For the AUPR and AUC values in the U1 split, the MSEDDI algorithm exhibited marginally better performance compared to KITE-DDI. This occurs as the positive and negative threshold hyperparameter values are set automatically for the proposed algorithm resulting in a smaller gap between them. 

The results obtained from running experiments on Dataset 2 show similar trends with KITE-DDI outperforming the other methods in the different performance metrices for both the U1 and U2 tasks. The values for most of the metrices are greater in general for Dataset 2 compared to Dataset 1. This happens as the first Dataset is smaller than the second, resulting in a larger training set size, making the former a more challenging test case.  

Also, it could be seen that the gap between the performance scores between KITE-DDI and the benchmark models are greater for the U2 task compared to the U1 task. This shows that the proposed model is able to perform better compared to the other algorithms when there is significant difference between the contents of the training and the test set proving that KITE-DDI is better at generalizing in more challenging circumstances. Another pattern that emerges from these results is the better outperformance of KITE-DDI compared to MSEDDI in the first Dataset than Dataset 2 indicating that the proposed model performs better with limited training data compared to the previous algorithms.

\begin{table}[!t]
\centering
\renewcommand{\arraystretch}{1.6}
\caption{\textbf{Ablation study comparing the performance of different developed models with the primary benchmark method, MSEDDI}}
\label{t_result2}
\setlength{\tabcolsep}{2pt}
\begin{tabular}{|c||c|c|c|c|c|c|c|c|c|}
\hline
\textbf{Models} & \begin{tabular}[c]{@{}c@{}}\textbf{Dataset}\\ \textbf{split}\end{tabular} & \textbf{Accuracy} & \begin{tabular}[c]{@{}c@{}}\textbf{F1 Score}\\ \textbf{(weighted)}\end{tabular} & \begin{tabular}[c]{@{}c@{}}\textbf{F1 Score}\\ \textbf{(macro)}\end{tabular} & \textbf{MCC} & \begin{tabular}[c]{@{}c@{}}\textbf{Precision}\\ \textbf{(Weighted)}\end{tabular} & \begin{tabular}[c]{@{}c@{}}\textbf{Precision}\\ \textbf{(macro)}\end{tabular} & \begin{tabular}[c]{@{}c@{}}\textbf{Recall}\\ \textbf{(weighted)}\end{tabular} & \begin{tabular}[c]{@{}c@{}}\textbf{Recall}\\ \textbf{(macro)}\end{tabular} \\ \hline \hline
\multirow{3}{*}{\textbf{BERT\_base}} & Eval & 68.61\% & 0.6778 & 0.4092 & 0.6184 & 0.6799 & 0.4690 & 0.6861 & 0.3900 \\ \cline{2-10} 
 & U2 & 25.28\% & 0.2338 & 0.0428 & 0.0564 & 0.2429 & 0.0419 & 0.2528 & 0.0595 \\ \cline{2-10} 
 & U1 & 46.29\% & 0.4477 & 0.2575 & 0.3365 & 0.4590 & 0.3079 & 0.4629 & 0.2603 \\ \hline \hline
\multirow{3}{*}{\textbf{BERT\_small}} & Eval & 66.01\% & 0.6588 & 0.3715 & 0.5928 & 0.6637 & 0.3917 & 0.6601 & 0.3778 \\ \cline{2-10} 
 & U2 & 23.15\% & 0.2388 & 0.0292 & 0.0701 & 0.3008 & 0.0446 & 0.2315 & 0.0285 \\ \cline{2-10} 
 & U1 & 44.33\% & 0.4441 & 0.2220 & 0.3297 & 0.4667 & 0.2299 & 0.4433 & 0.2588 \\ \hline \hline
\multirow{3}{*}{\textbf{BERT\_large}} & Eval & 71.58\% & 0.7097 & 0.4541 & 0.6560 & 0.7154 & 0.5471 & 0.7157 & 0.4307 \\ \cline{2-10} 
 & U2 & 26.30\% & 0.2478 & 0.0329 & 0.0811 & 0.2771 & 0.0506 & 0.2630 & 0.0310 \\ \cline{2-10} 
 & U1 & 47.55\% & 0.4660 & 0.2566 & 0.3560 & 0.4734 & 0.2925 & 0.4755 & 0.2726 \\ \hline \hline
\multirow{3}{*}{\textbf{TRFM\_conv}} & Eval & 77.02\% & 0.7636 & 0.5447 & 0.7221 & 0.7716 & 0.6515 & 0.7702 & 0.5062 \\ \cline{2-10} 
 & U2 & 30.09\% & 0.2650 & 0.0615 & 0.0904 & 0.2888 & 0.0900 & 0.3009 & 0.0684 \\ \cline{2-10} 
 & U1 & 51.09\% & 0.4915 & 0.2904 & 0.3935 & 0.5177 & 0.3467 & 0.5110 & 0.2985 \\ \hline \hline
\multirow{3}{*}{\begin{tabular}[c]{@{}c@{}}\textbf{TRFM}\_\textbf{conv}\\ \textbf{(linear Encoder)}\end{tabular}} & Eval & 82.59\% & 0.8238 & 0.6455 & 0.7904 & 0.8274 & 0.6852 & 0.8259 & 0.6486 \\ \cline{2-10} 
 & U2 & 30.83\% & 0.2823 & 0.0465 & 0.1246 & 0.2832 & 0.0533 & 0.3083 & 0.0490 \\ \cline{2-10} 
 & U1 & 50.58\% & 0.4936 & 0.3135 & 0.3938 & 0.5074 & 0.3345 & 0.5058 & 0.3572 \\ \hline \hline
\multirow{3}{*}{\textbf{TRFM\_pretrained}} & Eval & 87.43\% & 0.8731 & 0.7044 & 0.8488 & 0.8761 & \textbf{0.7354} & 0.8743 & \textbf{0.6956} \\ \cline{2-10} 
 & U2 & 41.11\% & 0.3803 & \textbf{0.1199} & 0.2437 & 0.3919 & 0.1016 & 0.4111 & \textbf{0.1167} \\ \cline{2-10} 
 & U1 & 58.92\% & 0.5784 & 0.4409 & 0.4939 & 0.5920 & 0.4741 & 0.5892 & 0.4403 \\ \hline \hline
\multirow{3}{*}{\textbf{MSEDDI}} & Eval & 87.47\% & 0.8728 & 0.6287 & 0.8489 & 0.8737 & 0.6330 & 0.8747 & 0.6151 \\ \cline{2-10} 
 & U2 & 44.33\% & 0.4087 & 0.0844 & 0.3009 & 0.4119 & \textbf{0.1027} & 0.4433 & 0.0845 \\ \cline{2-10} 
 & U1 & 65.11\% & 0.6368 & 0.4128 & 0.5688 & 0.6501 & 0.5018 & 0.6511 & 0.4049 \\ \hline \hline
\multirow{3}{*}{\textbf{KITE-DDI}} & Eval & \textbf{89.01\%} & \textbf{0.8883} & \textbf{0.6744} & \textbf{0.8678} & \textbf{0.8896} & 0.7134 & \textbf{0.8900} & 0.6562 \\ \cline{2-10} 
 & U2 & \textbf{51.30\%} & \textbf{0.4698} & 0.0657 & \textbf{0.3679} & \textbf{0.4737} & 0.0860 & \textbf{0.5130} & 0.0614 \\ \cline{2-10} 
 & U1 & \textbf{67.21\%} & \textbf{0.6628} & \textbf{0.5118} & \textbf{0.5970} & \textbf{0.6701} & \textbf{0.5580} & \textbf{0.6721} & \textbf{0.5331} \\ \hline
\end{tabular}
\end{table}

Next, the ablation study for the different versions of the proposed algorithm and the contribution of the different modules in the model are shown in Table \ref{t_result2}. Experiments with six different variants of the proposed model are carried out and the results examined, where architectural modifications are progressively made to reflect the contribution of each of these modification on the overall performance of the model. The different evaluation metrics that are used for comparison in these experiments include accuracy, macro F1 score, weighted F1 score, Mathews correlation coefficient (MCC) score, weighted precision, macro precision, weighted recall and macro recall. The results from the principal benchmark model, MSEDDI are also included in this table for comparison. The description of each of the ablation variants are given below: 

\begin{itemize}
  \item \textbf{BERT\_base:} This variant is based on the basic BERT model which consists of just an encoder with 8 encoder layers each of which consists of 6 attention heads. The dimension of the model is 256 with a feedforward network dimension of 1024. The drug pairs in each datapoint are tokenized and inserted into the encoder via the non-positional and positional encoders. The first element in the input sequence is set as the CLS token and its corresponding output goes into a classification header to predict the DDI labels. 
  \item \textbf{BERT\_small:} This variant is a miniaturized version of the BERT\_base model with the same number of encoders and attention heads but with a feed forward network dimension of 256 instead of 1024. 
  \item \textbf{BERT\_large:} This model consists of 12 encoder layers stacked one after the other with each consisting of 12 multiheaded attention blocks. The dimension of the model is 768 and the feed forward network has a length of 3072. 
  \item \textbf{TRFM\_conv:} In this variant, a convolutional module is inserted after the encoder layers to collect and process the information coming from the latent vector of the output sequences. It is made up of convolutional layers, batch normalization, RELU activation and pooling layers with skip connections to aid in the propagation of input data. The output from the conv layer is then fed into a classification header to generate the DDI predicitons.
  \item \textbf{TRFM\_conv (linear\_Encoder)}: This model is a modification of the previous one with a linear positional encoder used instead of the sinusoidal one which is the conventional transformer positional encoding technique.
  \item \textbf{TRFM\_pretrained:} In this model, the transformer encoder architecture is pretrained on an unlabeled dataset of drug SMILEs before finetuning it on the DDI Datasets. 
  \item \textbf{KITE-DDI:} This is the final model where the DRKG drug embeddings are integrated into the model to increase the generalizability of the model on inductive tasks. 
\end{itemize}

In order to compare the contribution of the different variants, their performance using various evaluation matrices are given in Table \ref{t_result2} for the validation, U1 and U2 test sets. The results show gradual increase of the performance of the model as successive modules are added into it. The first three models are modified versions of the BERT\cite{BERT} architecture which consists of a classification output token to make the output prediction. The bert\_large\_reg model has been the most successful in terms of most metrics among these three. 

The next model, Trfm\_conv, consists of a convolutional module after the encoder block to process information from the output sequences. This modification demonstrates a notable enhancement in the model's performance, while also being more efficient and demanding fewer computational resources than the Bert\_large\_reg model. Next, the sinusoidal encoder is replaced by a linear one to streamline the encoding process and finally, the model is pretrained on an unlabeled dataset of drug SMILES to enhance the accuracy and performance of the model even further. Lastly, the drug embeddings extracted from a biomedical knowledge graph are incorporated into the transformer architecture with the help of self-attention modules to develop the final KITE-DDI model.

\begin{figure}[h]
  \centering
  % \fbox{\rule[-.5cm]{4cm}{4cm} \rule[-.5cm]{4cm}{0cm}}
  \includegraphics[width=\linewidth]{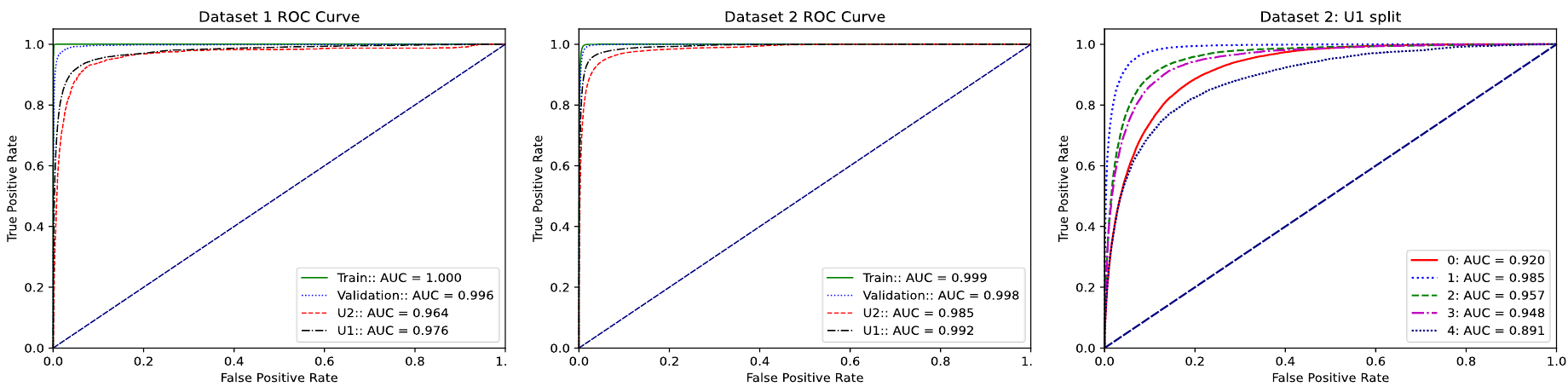}
  \caption{Left: Averaged ROC curve of Dataset 1 for the train, validation, U1 and U2 splits. Center: Averaged ROC curve of Dataset 2 for the train, validation, U1 and U2 splits. Left: Individual ROC curve for the 5 most populous classes in Dataset 2 for the U1 split.}
  % \Description{ROC curves}
  \label{Fig_ROC}
\end{figure}

% \Figure[t!](topskip=0pt, botskip=0pt, midskip=0pt)[width=1\linewidth]{Fig_ROC_cropped1.pdf}
% { \textbf{Left: Averaged ROC curve of Dataset 1 for the train, validation, U1 and U2 splits. Center: Averaged ROC curve of Dataset 2 for the train, validation, U1 and U2 splits. Left: Individual ROC curve for the 5 most populous classes in Dataset 2 for the U1 split}\label{Fig_ROC}}

The Receiver Operating Characteristic (ROC) curve of the proposed model for Datasets 1 and 2 are given in Fig. \ref{Fig_ROC}. The ROC curve plots False positive rate vs True positive rate to show how well the algorithm performs at differentiating between the positive and negative samples. In the most ideal situation, the ROC curve should take the shape of an inverted ‘L’ which maximizes the area under the curve. The first and second plot shows the ROC curves for the Training, validation, U1 and U2 split results of the Kite-DDI model for Dataset 1 and 2 respectively. A unit gradient line is also drawn to represent the output for a perfectly random system. The results show that the algorithm is doing very well in differentiating between the positive and negative samples in both Dataset 1 and 2. The area under the curve which is represented by the parameter, AUC, is also above 0.95 for both the U1 and U2 inductive splits for Datasets 1 and 2. 

Lastly, the third plot shows the individual ROC curves for the top 5 most populous classes on Dataset 2. It shows that the proposed algorithm is better at discerning some classes compared to others. This happens as some of the DDI events have longer, unique and more prominent markers to identify them compared to others which are more difficult to differentiate. However, even the more difficult classes have AUC of around 0.9 or more which shows that the algorithm is still able to differentiate between them with adequate accuracy.

\begin{figure}[h]
  \centering
  % \fbox{\rule[-.5cm]{4cm}{4cm} \rule[-.5cm]{4cm}{0cm}}
  \includegraphics[width=0.6\linewidth]{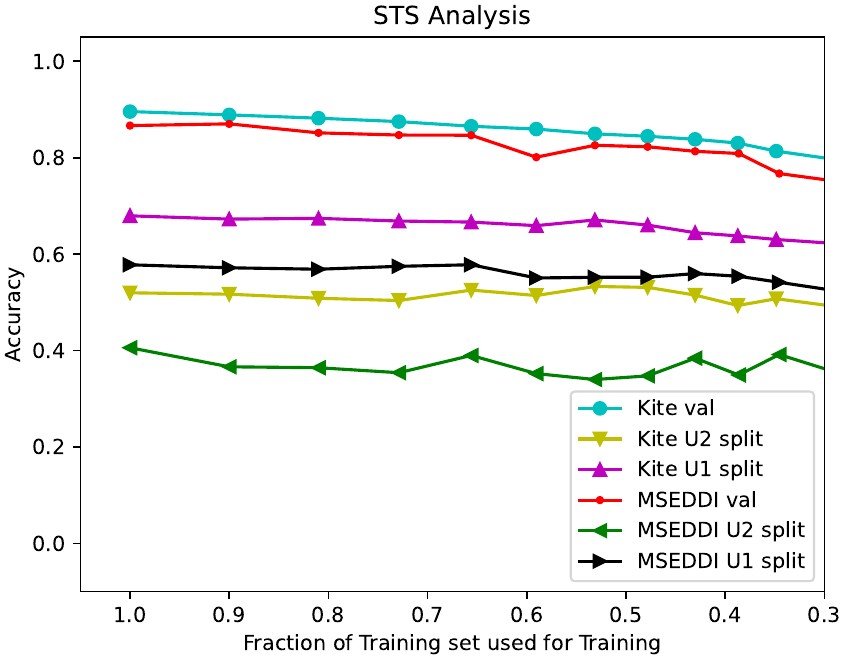}
  \caption{STS analysis of KITE-DDI model on Dataset 1 for the validation, U1 and U2 splits including comparison with the MSEDDI model.}
  % \Description{ROC curves}
  \label{Fig_STS}
\end{figure}

% \Figure[t!](topskip=0pt, botskip=0pt, midskip=0pt)[width=0.99\linewidth]{Fig_STS_analysis.pdf}
% { \textbf{STS analysis of KITE-DDI model on Dataset 1 for the validation, U1 and U2 splits including comparison with the MSEDDI model.}\label{Fig_STS}}

\subsection{STS Analysis}
In order to test the performance of the model at low data scenarios, a Shrinking Training Set (STS) analysis is carried out on the proposed algorithm and comparisons have been made with the main benchmark model and the results illustrated in Fig. \ref{Fig_STS}. In each step of the process, the dataset used to train the model is shrunk by 10\% and the new accuracy values are computed. The size of the validation and the U1 and U2 splits are kept the same during the process. Dataset 1 is used as the starting point and all classes which have less than 5 samples in the training set are discarded to make sure that all the datasets have the same number of classes. A stratified split is then done on the training set to split it into two segments of 90\% and 10\%. The bigger segment is kept and the smaller one discarded. This process is repeated until the training set size falls to around 7\% of its initial size. 

The Validation, U1 and U2 accuracies are plotted in Fig. \ref{Fig_STS} for different training set sizes along with the corresponding values for the MSEDDI model. The results show that the proposed architecture maintains its accuracy consistently even at very small training set sizes and manages to keep its performance advantage over the primary benchmark model, MSEDDI, in all the different training set sizes. Also, there are limited fluctuations in accuracy while varying the training set which shows the stability and robustness of the proposed architecture.

\begin{figure}[h]
  \centering
  % \fbox{\rule[-.5cm]{4cm}{4cm} \rule[-.5cm]{4cm}{0cm}}
  \includegraphics[width=0.6\linewidth]{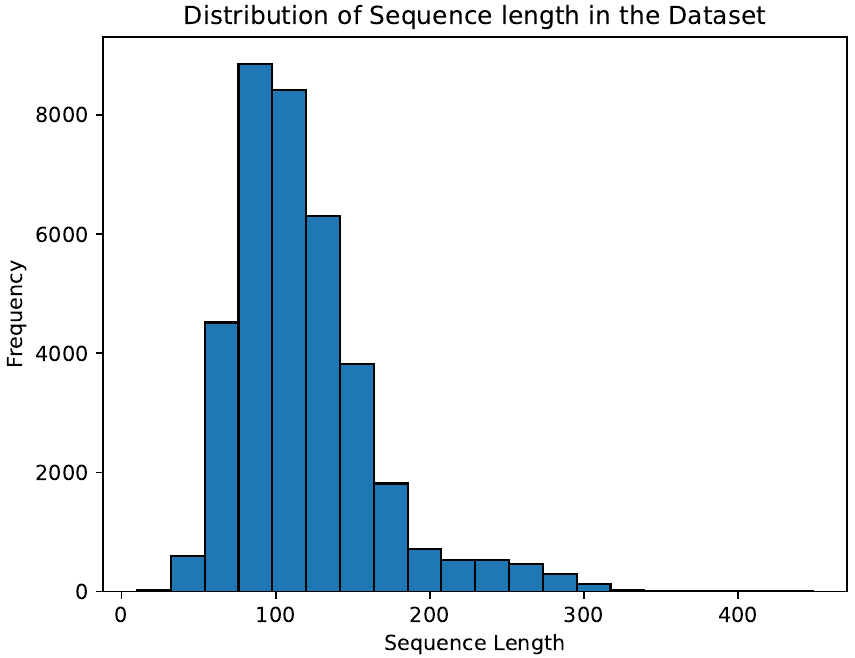}
  \caption{Histogram showing the frequency distribution of SMILES with different lengths in Dataset 1.}
  % \Description{ROC curves}
  \label{Fig_seq1}
\end{figure}

% \Figure[t!](topskip=0pt, botskip=0pt, midskip=0pt)[width=0.99\linewidth]{Fig_seq_dist.pdf}
% { \textbf{Histogram showing the frequency distribution of SMILES with different lengths in Dataset 1.}\label{Fig_seq1}}

\begin{figure}[h]
  \centering
  % \fbox{\rule[-.5cm]{4cm}{4cm} \rule[-.5cm]{4cm}{0cm}}
  \includegraphics[width=0.6\linewidth]{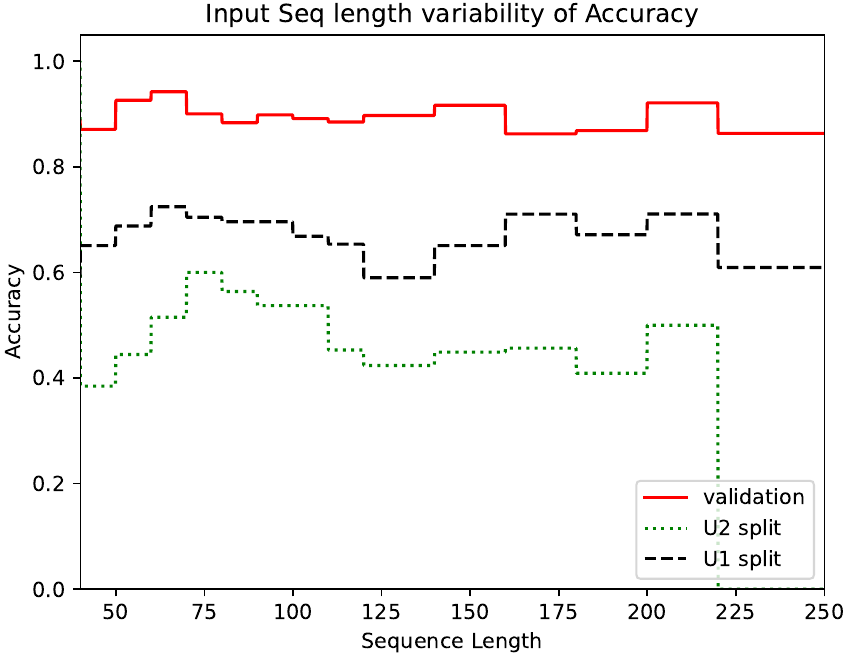}
  \caption{Variability of accuracy with input sequence length for the KITE-DDI model on Dataset 1 for the validation, U1 and U2 splits.}
  % \Description{ROC curves}
  \label{Fig_seq2}
\end{figure}

% \Figure[t!](topskip=0pt, botskip=0pt, midskip=0pt)[width=0.99\linewidth]{Fig_seq_len_analysis_dataset1.pdf}
% { \textbf{Variability of accuracy with input sequence length for the KITE-DDI model on Dataset 1 for the validation, U1 and U2 splits.}\label{Fig_seq2}}

\subsection{Sequence Length Analysis} 
This section analyzes the variability of the model performance and its dependencies on the input sequence size. As a sequence model based on the transformer architecture, KITE-DDI is able to take input sequences of any sizes. The distribution of input sequence length in the dataset is given in Fig. \ref{Fig_seq1}. Most of the drug compounds have a SMILE length from 55 to 200 tokens with a peak of 100 bases and a narrow tail at the end. The accuracy of the algorithm at different sequence length is plotted in Fig. \ref{Fig_seq2}. Here, the datapoints are distributed in different bins based on their token lengths and the mean accuracy for each bin is represented by the graph. It shows that the proposed algorithm can hold consistent accuracy with minimal fluctuations for different input lengths and its performance is not dependent on the length of the input sequences. Maintaining similar accuracy even with longer input sequences indicates that the model can understand and learn the long-term patterns and dependencies within the input and can generalize well to a wide range of varying sequence lengths. However, the accuracy of the U2 split does drop to zero after an input length of 225. This happens as the U2 dataset is quite small and it only contains two samples that are longer than 225 tokens in the dataset leading to the result being irrelevant beyond that point.

\section{Conclusion}
In this work, we have implemented a transformer based lightweight end-to-end deep learning system, KITE-DDI, for predicting DDI events with higher precision and accuracy compared to other similar state-of-the-art methods. The algorithm integrates the information from Biomedical Knowledge graphs with SMILES representations of drug compounds with the help of a transformer encoder, a convolutional module, and a self-attention block to make the predictions. The model's performance was evaluated on two prominent benchmark datasets using several metrics. The results showed that KITE-DDI performed significantly better compared to other contemporary methods specially on inductive datasets, while at the same time not dependant on external heuristic algorithms for feature extraction. The model is also easy to use, requiring very few hyperparameters to train effectively and shows consistent performance even at very low training set sizes. Furthermore, the system is able to automatically maintain balance between precision and recall without having to tune any additional hyperparameter and its accuracy is independent of the input sequence length making it robust and generalizable.

Although there have been significant work and scientific studies conducted in the field of DDI predictions in recent times, there are still many areas of improvement in the domain. Some of these potential avenues of exploration and investigation includes expanding DDI prediction models to other non small molecule drugs like RNA based medications. Additional work is also required in collecting more samples for rarer DDI events to create a less skewed dataset which could be used to train more efficient and comprehensive DDI prediction models. Hence, there is a considerable amount of work that must be undertaken in this field, encompassing a multitude of possible avenues for exploration and investigation.

\bibliographystyle{unsrt}  
\bibliography{main}

\end{document}